\documentclass[11pt, a4paper, twocolumn, copyright]{google}

\usepackage[authoryear, sort&compress, round]{natbib}
\makeatletter
\renewcommand{\absfont}{\normalfont\linespread{1.2}\fontsize{11}{12}\selectfont}
\makeatother
\definecolor{linkblue}{RGB}{0,0,139}      % 深蓝
\definecolor{navy}{RGB}{0,0,128}          % 海军蓝
\definecolor{royalblue}{RGB}{65,105,225}  % 皇家蓝
\definecolor{steelblue}{RGB}{70,130,180}  % 钢蓝
\definecolor{dodgerblue}{RGB}{30,144,255} % 道奇蓝
\definecolor{mediumblue}{RGB}{0,0,205}    % 中蓝
\definecolor{darkslateblue}{RGB}{72,61,139} % 深岩蓝
\usepackage[colorlinks = true,
            linkcolor = linkblue,
            urlcolor  = linkblue,
            citecolor = linkblue, % brown
            anchorcolor = blue]{hyperref}

\usepackage[misc]{ifsym}
\usepackage{minitoc}
\usepackage{cleveref} 
\usepackage{subcaption}
\usepackage{booktabs}
\usepackage{amsmath}
\usepackage{mathtools}
\usepackage{amssymb}
\usepackage{graphicx}   % for \resizebox
\usepackage{cuted}
\usepackage{booktabs}   % for \toprule etc. \bottomrule
\usepackage{algorithm}
\usepackage{tcolorbox}
\usepackage{verbatim}   % for comment environment
\usepackage{makecell}   % line breaks inside cells
\usepackage{array}      % better column
\usepackage{multirow}
\usepackage{xurl}
\usepackage{amsthm}
\usepackage{booktabs}
\usepackage{adjustbox}
\usepackage{caption}

\usepackage{amssymb}% http://ctan.org/pkg/amssymb
\usepackage{pifont}% http://ctan.org/pkg/pifont
\usepackage{algorithm}% http://ctan.org/pkg/algorithm
\usepackage{algpseudocode}% http://ctan.org/pkg/algorithmicx

\usepackage{listings}

\tcbuselibrary{listings,skins,breakable}

\usepackage{tikz}
\usetikzlibrary{
    matrix,
    arrows.meta,
    positioning,
    shapes.geometric,
    fit,
    backgrounds
}

\usepackage{pdflscape}
\usepackage{afterpage}

\newcommand{\app}{\textsc{ASTRA}}
\newcommand{\whisper}{\emph{Whisper}}

\makeatletter

\newcommand{\appendixtableofcontents}{%
  \begingroup
  \renewcommand{\contentsname}{Appendix Contents}%
  \@starttoc{atoc}%
  \endgroup
}

\newcommand{\startappendixtoc}{%
  \let\oldaddcontentsline\addcontentsline
  \renewcommand{\addcontentsline}[3]{%
    \def\@tempa{##1}\def\@tempb{toc}%
    \ifx\@tempa\@tempb
      \oldaddcontentsline{atoc}{##2}{##3}%
    \else
      \oldaddcontentsline{##1}{##2}{##3}%
    \fi
  }%
}
\makeatother

\title{\app{}: A Scalable Next-Generation ATCO Training Simulator with Autonomous Simpilots}

\correspondingauthor{lim\_yong\_zhi1@defence.gov.sg}

\renewcommand{\today}{}
\author[1]{Ethan Chew\textsuperscript{*}}
\author[1]{Enjia Wu\textsuperscript{*}}
\author[1]{Iruss Eng}
\author[1]{Ian Lim}
\author[1]{Ranen Sim}
\author[1]{Brandon Koh}
\author[1]{Kaleb Nim}
\author[1]{Caden Toh}
\author[1]{Wei Dong Soin}
\author[1]{Darius Koh}
\author[1]{Galen Tay}
\author[1]{Prannaya Gupta}
\author[1]{Jonathan Koong}
\author[1 \ \Letter]{Yong Zhi Lim}

\affil[*]{Equal Contribution}
\affil[1]{Air Emerging Technologies High-Speed Experimentations and Research (AETHER), RSAF Agile Innovation Digital (RAiD), Republic of Singapore Air Force, Singapore}
\affil[ \Letter]{Corresponding Author}

\begin{abstract}
Air Traffic Control Operators (ATCOs) are vital in ensuring the safe, orderly, and efficient flow of air traffic, yet training capacity is constrained by reliance on specialized human trainers known as \textit{simpilots}, who must role-play both pilots and ATCOs in a simulated airspace. Existing automated solutions rely on Western-centric speech models that perform poorly in Singaporean operational contexts, with off-the-shelf systems exhibiting Word Error Rates (WER) of up to 107.80\% on Singaporean-accented aviation speech. We introduce \app{}, an end-to-end training simulator that automates these \textit{simpilot} roles through a pipeline that transcribes ATCO speech, interprets instructions, and generates appropriate pilot and ATCO responses using locally adapted voice models. Our fine-tuned Automatic Speech Recognition (ASR) pipeline reduces WER to 23.45\%, substantially outperforming existing approaches in this domain. Beyond traffic simulation, \app{} incorporates an AI-assisted performance evaluation framework that assesses trainee radiotelephony communications across accuracy, brevity, and completeness, achieving post-optimization scores of 91.7\%, 88.2\%, and 86.9\%, respectively. Built on open-source foundations such as DSPy and Unsloth, this approach enables scalable, standardized ATCO assessment while reducing instructor workload.

\end{abstract}

\begin{document}

\maketitle

\begin{strip}
    \centering
    \includegraphics[width=\linewidth]{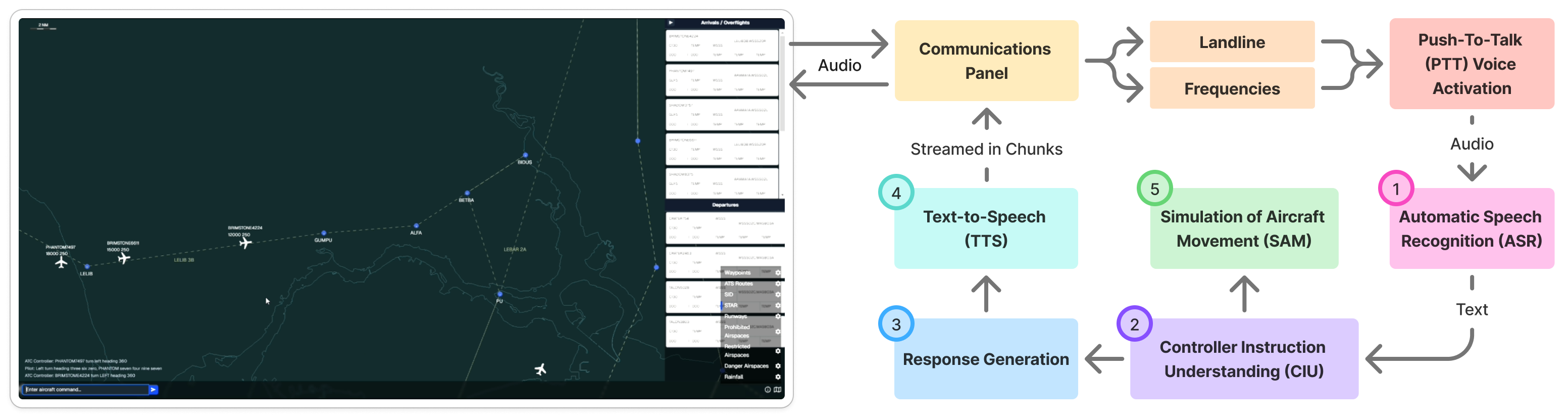}
    \captionof{figure}{Management of Communications between ATCO Trainee and ASTRA Training Simulator}
    \label{fig:comms-stack}
    \vspace{15pt}
\end{strip}

\section{Introduction}
Air Traffic Control Operators (ATCOs) play a critical role in ensuring the safe, orderly, and efficient flow of air traffic in increasingly congested skies. 
% This ensures the orderly, efficient, and timely flow of aircraft both on the ground and in the air, enabling smooth operations and directly contributing to mission success. % ethan: not sure if we wanna keep this
As aviation recovers rapidly from COVID-19, there is a critical shortage of ATCOs, increasing safety risks and limiting capacity~\citep{understaffed-cbs}.
%problem statement
Current ATCO training systems heavily rely on specialized human trainers known as simulation-pilots (or \textit{"simpilots"}) role-playing both aircraft pilots (or \textit{"pseudo pilots"}) and other ATCOs (or \textit{"ghost controllers"}) to coordinate with the trainee to create a realistic training environment~\citep{faa_simpilots}.

% This paradigm struggles with scalability; the complement of instructor and \textit{simpilots} limits the number of trainees that can be accommodated at one time.
% %solution
% Automating the role of \textit{simpilots} is essential to increase ATCO training capacity, a challenge that an autonomous system is well-equipped to tackle~\cite{brudnicki2005application}.
This paradigm struggles with scalability: the required complement of instructors and simpilots limits training throughput and constrains practice to moments when qualified personnel are co-located and available. Automating the simpilot role is therefore essential to enable more flexible training, while providing standardized and objective assessment independent of human availability ~\citep{brudnicki2005application}. A further constraint is localization, where frontier models trained on American and British English fail on \textit{Singaporean}-accented speech and aviation terminology.

This work introduces \textbf{\app{}}, a training simulator that mimics the functionality of a \textit{simpilot}. \app{} implements an end-to-end speech-to-speech pipeline to respond to input commands from ATCO trainees with appropriate responses.

This pipeline consists of five major stages, which are modeled after~\citet{lin2021deep}:

\begin{enumerate}
    \item \textbf{Automatic Speech Recognition (ASR)}: transcribes ATCO trainee speech into textual commands.
    \item \textbf{Controller Instruction Understanding (CIU)}: parses textual commands and extracts structured target parameters (STPs).
    \item \textbf{Response Generation}: generates contextually appropriate replies from STPs from either the \textit{pseudo pilot} or \textit{ghost controller}.
    \item \textbf{Text-to-Speech (TTS)}: synthesizes audio based on responses generated by the Response Generation module, which is then streamed to the ATCO trainee.
    \item \textbf{Simulation of Aircraft Movement (SAM)}: utilizes STPs to reflect changes in the positions of the aircraft in the simulation environment.
\end{enumerate}

Beyond \textit{simpilot} automation, \app{} addresses a second gap in existing training systems: the lack of objective, scalable performance assessment. \app{} therefore incorporates an AI-assisted performance evaluation framework that scores trainee's radiotelephony across accuracy, brevity, and completeness, providing automated feedback that would have previously required an experienced instructor to deliver.

The rest of this paper is structured as follows: \Cref{existing-works} reviews relevant literature across the key components of automated ATCO training systems, outlining the current state of the art. \Cref{technical_approach} describes the design and implementation of \app{}, including the simulation environment and end-to-end speech pipeline. \Cref{results} presents experimental evaluations of the system, covering ASR, TTS, and communication assessment performance. Finally, \Cref{futurework} discusses the limitations of the current approach and outlines directions for future work, before concluding in \Cref{conclusion}.

%%%%%%%%%%%%%%%%%%%%%%%%%%%%%%%%%%%%%%%%%%%%%%%%%%%%%%%%%%%%%%%%%%%%%%%%%%%%%%%%%%%%%%%%%%%%%%%%%%%%%%%%%%%%%%%%%%%%%%%%%%
% EXISTING WORK %
%%%%%%%%%%%%%%%%%%%%%%%%%%%%%%%%%%%%%%%%%%%%%%%%%%%%%%%%%%%%%%%%%%%%%%%%%%%%%%%%%%%%%%%%%%%%%%%%%%%%%%%%%%%%%%%%%%%%%%%%%%
\section{Existing Work} \label{existing-works}

Our extensive review of works related to \textit{simpilots} is decomposed into four sub-modules: \textit{ASR}, \textit{CIU}, \textit{Response Generation}, and \textit{TTS}.

\subsection{Automatic Speech Recognition (ASR)}
Modern ASR systems such as \whisper{}~\citep{radford2023robust} have made significant advancements, resulting in broader applicability across domains. Despite these advancements, these frontier ASR models tend to struggle with the following two issues: transcribing Singaporean-accented speech and accurately recognizing domain-specific terminology.

\subsubsection{Accent Robustness in ASR Models}

Conventional ASR models perform poorly on Singaporean-accented English due to limited speech corpora available for training with these accents.

\citet{he2024MERaLiON} and \citet{wang2025advancing} proposed \emph{MERaLiON}, an audio foundation model capable of transcribing local accented speech more accurately as compared to existing frontier audio foundation models, and introduced the Multitask National Speech Corpus (MNSC), a large-scale corpus for Singaporean-accented transcription.

\subsubsection{Domain Terminology in Aviation ASR}
Conventional ASR models struggle with accurately recognizing specific terminology used in radiotelephony communications. To address this, many works explore fine-tuning to adapt the weights of such models (e.g. \whisper{}) to domain-specific corpora.

\citet{van2024whisper} presented \emph{WhisperATC}, a family of models fine-tuned on ATCOSIM~\citep{hofbauer2008atcosim} and ATCO2~\citep{zuluaga2022atco2}. These models achieve a Word Error Rate (WER) of $16.74\%$ for ATCO2 and $1.19\%$ for ATCOSIM, as compared to base \whisper{} which achieved $24.03\%$ and $16.74\%$ respectively.

\subsection{Controller Instruction Understanding}

\citet{prasad2022speech} and \citet{zuluaga2023virtual} trained a named entity recognition (NER) model based on BERT~\citep{devlin2019bert}, which attempts to break down an ASR command into three key fields, 1) \textit{callsign}, 2) \textit{command} and 3) \textit{value}.
\citet{jiang2024slkir} proposed Small Sample Learning for Key Information Recognition (SLKIR), an end-to-end deep learning framework for information extraction from Chinese ATC commands.

\subsection{Response Generation}

\citet{lin2021deep} proposes a pilot repetition generation method by training a \emph{Seq2Seq} model to generate pilot readbacks by reordering and preserving key elements of the ATCO instruction, such as aircraft callsign and command parameters. Their bidirectional long-short term memory network with attention helps capture ATC-specific structures and cases where only partial readback is required.

\subsection{Text-to Speech (TTS)}

Neural TTS has rapidly advanced with multilingual and voice-cloning models such as \emph{VITS}~\citep{kim2021conditional} and \emph{XTTS}~\citep{casanova2024xtts}, enabling natural and intelligible speech for aviation training.

\citet{10765121} introduced a TTS system for ATCO and pilot speech, fine-tuning \emph{VITS} and \emph{XTTS} on ATCOSIM and multilingual pilot-speech datasets. Across 4100 subjective ratings, \emph{XTTS} outperformed \emph{VITS} in clarity, pronunciation, intonation, naturalness, and overall quality.

Nonetheless, current ATC-focused TTS systems still struggle with the following:
1) generating consistent Singaporean-accented speech,
2) producing accurate pronunciation of aviation-specific terms,
and 3) limited domain-appropriate evaluation metrics.

\subsubsection{Accent Robustness in TTS Models}
Most aviation TTS systems struggle with underrepresented accents because training data is dominated by American or British English, limiting local phonetic and prosodic modeling. Insufficient regional data leads to accent drift and unstable prosody~\citep{10.1109/TASLP.2024.3363414}.
Zero and few-shot models like \emph{XTTS} mitigate this using multilingual representations for low-resource accent transfer.

\subsubsection{Domain Terminology in Aviation TTS}
Accurate aviation radiotelephony requires TTS models to pronounce domain-specific terms often missing from general corpora, causing mispronunciations and irregular pacing. Low-frequency or unseen tokens produce unstable pronunciation, and reliable output needs specialized lexicons or pronunciation dictionaries~\citep{ttsdomainadapt}. 

\citet{hu2019domain} augmented training with synthetic word–pronunciation pairs and phoneme improvements, reducing errors on unseen terms, while adding domain-specific lexicons and rules to stabilize output.

\subsubsection{Limitations of TTS Evaluation Methods}

TTS systems lack reliable, domain-specific evaluation metrics. Mean Opinion Scores (MOS) are subjective and difficult to attain, while Mel Cepstral Distortion does not capture domain phraseology or radiotelephony timing. ASR-based intelligibility is also unreliable across mismatched domains~\citep{salesky2021assessing}, and without a Singaporean-accented aviation ASR, scores largely reflect ASR bias rather than TTS quality. Recent work explores automated evaluators like Large Language Model (LLM)-based scoring~\citep{wang2025enabling} and ASR-ensemble methods combining ASR confidence, pronunciation, and acoustic similarity~\citep{kirk2025mos}, but both still require domain-matched or calibrated models.

\subsection{AI-Assisted Radiotelephony Performance Evaluation}

Current evaluation methodologies rely on experienced instructors manually reviewing trainee communications, introducing variability and limiting throughput.

\citet{aldridge2025identifying} highlight that objective and continuous measurement of ATC performance remains relatively underexplored, emphasizing the need for structured and quantifiable evaluation methods. 

Early work by~\citet{brudnicki2005application} introduced Intelligent Tutoring Systems (ITS) to support structured evaluation in ATC training. ITS frameworks define three key components: 1) an \textit{Expert Model}, representing expected performance, 2) a \textit{Student Model}, capturing observed trainee behavior, and 3) an \textit{Instructor Model}, supporting feedback and after-action-review (AAR). However, practical adoption remains limited, with evaluation processes still largely reliant on manual interpretation.

To improve objectivity, rule-based evaluation methods have been explored.~\citet{wu2020rulebased} demonstrate that predefined scoring rules can provide consistent and interpretable assessments in ATC simulation environments. Nevertheless, such approaches remain rigid and are unable to effectively capture linguistic variation and contextual intent in radiotelephony communication.

Recent advances in large language models (LLMs) provide new opportunities to address these limitations.~\citet{chiang2023llm} show that LLMs can function as reliable evaluators when guided by structured prompts and rubrics, while~\citet{zhang2020bertscore} demonstrate that contextual embedding methods such as BERTScore enable semantic similarity evaluation beyond exact lexical matching. These developments motivate hybrid evaluation approaches that combine rule-based consistency with LLM-as-a-judge to support more robust and scalable radiotelephony performance assessment.

%%%%%%%%%%%%%%%%%%%%%%%%%%%%%%%%%%%%%%%%%%%%%%%%%%%%%%%%%%%%%%%%%%%%%%%%%%%%%%%%%%%%%%%%%%%%%%%%%%%%%%%%%%%%%%%%%%%%%%%%%%
% METHODOLOGY
%%%%%%%%%%%%%%%%%%%%%%%%%%%%%%%%%%%%%%%%%%%%%%%%%%%%%%%%%%%%%%%%%%%%%%%%%%%%%%%%%%%%%%%%%%%%%%%%%%%%%%%%%%%%%%%%%%%%%%%%%%
\section{Technical Approach} \label{technical_approach}

\app{} implements an internal realistic simulation engine with a context-appropriate user interface, an in-built scenario management system and an end-to-end speech pipeline.

\subsection{User Interface}

% To simulate traditional Air Traffic Control (ATC) interfaces, \app{} leverages the Unity game engine integrated with Cesium Ion for high-fidelity geospatial visualization. \app{} also incorporates a  communications panel designed to manage concurrent radio and landline interactions.

% Beyond the simulation environment, the \app{} User Interface serves as a centralized platform that allows instructors to configure training scenarios and monitor trainees.

To simulate traditional Air Traffic Control (ATC) interfaces with enhanced realism, \app{} leverages the Unity game engine integrated with Cesium Ion for high-fidelity geospatial visualization, including detailed 3D terrain and radar views that mirror real-world ATC displays. 
% \app{} also incorporates a communications panel designed to manage concurrent radio and landline interactions.

Beyond the simulation environment, the \app{} User Interface serves as a centralized platform that allows instructors to configure training scenarios and monitor trainees.

\subsection{Training Scenarios}

% Instructors can initiate self-contained simulation environments called \textit{sessions} for trainees. Within these \textit{sessions}, \textit{scenarios} can be created and loaded via a node-based customization tool.

% A \textit{scenario} is composed of \textit{aircraft profiles} that specify aircraft type, operational intent, assigned flight route and scenario-specific context. They utilize event-driven triggers to determine when aircraft are introduced, enabling autonomous management of aircraft behavior with dynamic sequencing that adapts to trainee actions and commands without further instructor intervention.

% Within the tool, each node represents an \textit{aircraft profile}. Connecting triggers across nodes establishes the sequence of actions that form a complete scenario. Instructors can define an action or event in a scenario by setting three parameters:
% \begin{itemize}
%     \item \textbf{Event Trigger}: A configurable rule attached to an aircraft in the scenario.
%     \item \textbf{Event Type}: The condition in the simulation to listen for (e.g. \texttt{AfterReportingWaypoint}). An \textit{Event Type} might require additional parameters to be filled in (e.g. Waypoint Name).
%     \item \textbf{Event Handler}: The action(s) that should run once the event type condition is met.
% \end{itemize}

\app{} supports two scenario modalities. In \textbf{Free-for-All} mode, pre-loaded aircraft create an open-ended environment where trainees manage traffic reactively, suited for advanced users.

In \textbf{Story Mode}, instructors design event-driven scenarios using a node-based tool. Scenarios consist of \textit{aircraft profiles} (aircraft type, intent, route, and context), with triggers controlling when and how aircraft are introduced, enabling dynamic sequencing that adapts to trainee actions without further intervention.

Within instructor-created \textit{sessions}, each node represents an \textit{aircraft profile}. Connecting triggers across nodes establishes the sequence of actions that form a complete scenario. Instructors can define an action or event in a scenario by setting three parameters:
\begin{itemize}
    \item \textbf{Event Trigger}: A configurable rule attached to an aircraft in the scenario.
    \item \textbf{Event Type}: The condition in the simulation to listen for (e.g. \texttt{AfterReportingWaypoint}). An \textit{Event Type} might require additional parameters to be filled in (e.g. Waypoint Name).
    \item \textbf{Event Handler}: The action(s) that should run once the event type condition is met.
\end{itemize}

\subsection{End-to-End Speech Pipeline}

\app{} implements an end-to-end speech pipeline to simulate the \textit{simpilot} communications, which is decomposed into five sub-modules: \textit{ASR}, \textit{CIU}, \textit{Response Generation}, \textit{TTS}, and \textit{SAM}.

\subsubsection{Automatic Speech Recognition (ASR)}

\app{} generically supports various frontier ASR models such as \whisper{} Small and Large v3 and \emph{Parakeet} TDT v3~\citep{sekoyan2025canary} to transcribe speech data from ATCO trainees. In addition, it also supports specialized ASR models such as \emph{MERaLiON}, a model trained on Singaporean-accented data, and \emph{WhisperATC} Small and Large v3, a family of models trained on ATC-specific data.
% \footnote{\href{https://hf.co/collections/aether-raid/asr-models-evaluated-for-astra}{List of models evaluated.}}

\begin{figure}[h!]
    \centering
    \includegraphics[width=\linewidth]{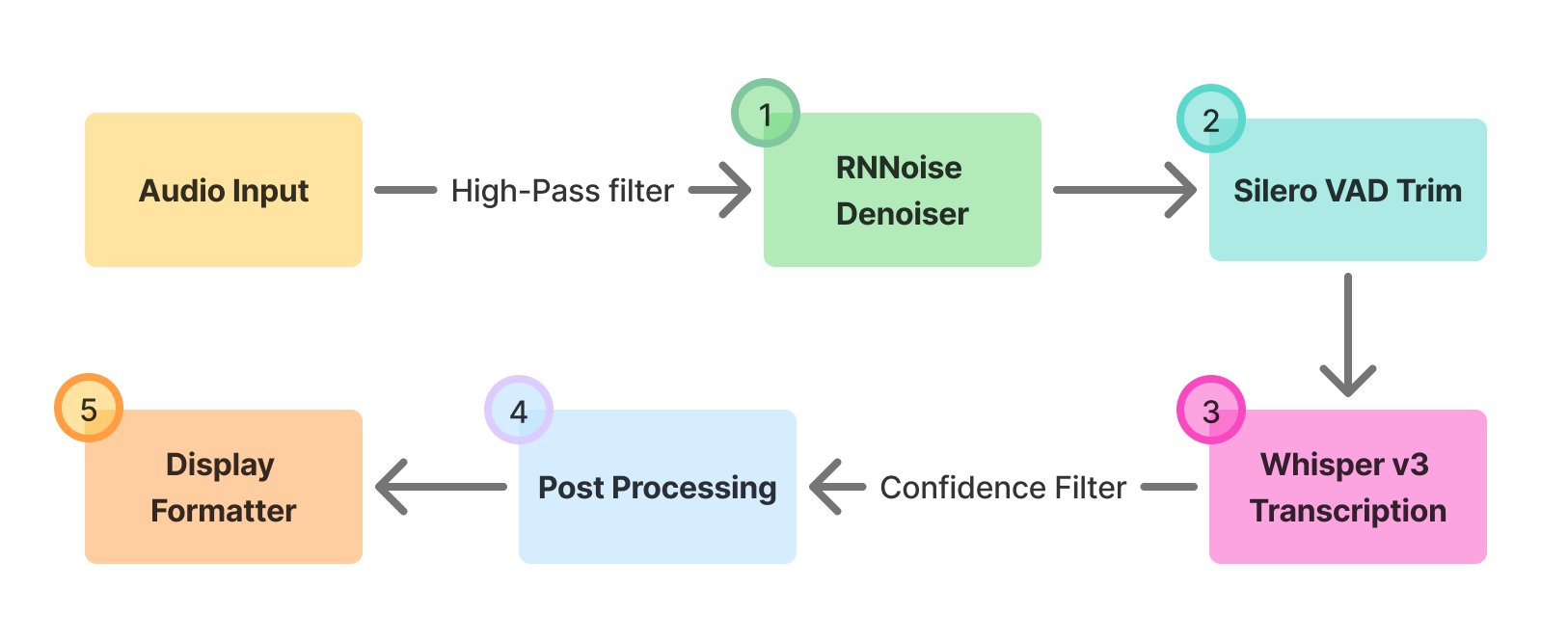}
    \caption{ASR inference pipeline}
    \label{fig:asr-pipeline}
\end{figure}

\app{} implements a multi-stage ASR inference pipeline optimized for noisy, accented ATC audio. The pipeline moves through four stages: audio preprocessing, voice activity detection, transcription, and post-processing (\Cref{fig:asr-pipeline}).

\begin{enumerate}[label=\alph*)]
    \item \textbf{Audio Preprocessing}: Raw audio is resampled to 16\,kHz, peak-normalized to $-1.0$\,dBFS, and passed through a high pass filter at 150\,Hz to suppress wind rumble and plosive artefacts common in VHF radio microphones. Noise reduction is then applied using \emph{RNNoise}~\citep{valin2018hybrid}, a recurrent neural network trained for real-time speech denoising. The audio is upsampled to 48 kHz via \emph{soxr}, for  processing, then downsampled back to 16 kHz.

    \item \textbf{Voice Activity Detection}: A \emph{Silero VAD}~\citep{silerovad} model trims the audio to speech-only regions before transcription. Using a high confidence threshold (0.85) with padding, the VAD aggressively removes leading and trailing silence, which is a known cause of \whisper{} hallucinations in quiet segments.

    \item \textbf{Transcription}: \app{} deploys a fine-tuned \whisper{} Large v3 model, converted to CTranslate2~\citep{ctranslate2} format via \emph{faster-whisper}~\citep{fasterwhisper} for reduced memory usage and faster inference. Beam search is biased with ~200 domain-specific hotwords covering Singapore locations, waypoints, ATC procedures, callsigns, and confusable terms. The ASR training corpus is mostly synthetic, generated using Chatterbox TTS and F5-TTS. 

    \item \textbf{Post-Processing}:
    To ensure correct pronunciation of Singaporean place names, TTS input uses simplified spoken forms (e.g., "leh-bar" for "lebar"), while the model is trained to output the standard spelling. Post-processing corrects edge cases where the model reproduces the spoken form instead.

    A rule-based formatter then converts the normalised text to a human readable format (e.g., \textit{``eagle one climb flight level two eight zero''} $\rightarrow$ \textit{``EAGLE 1 climb FL280''}). 
    %ethan: imo, we don't have to tell them about the "earlier" version
    %This replaced an earlier LLM-based formatter, reducing formatting latency from ${\sim}250$\,ms to ${<}1$\,ms while eliminating GPU memory usage.
    %ranen: ok!
\end{enumerate}

\subsubsection{Controller Instruction Understanding (CIU)}

\app{} implements a two-stage CIU processing pipeline for parsing ATC utterances, interpreting commands and extracting key information. To ensure low-latency processing, all command types are stored in a \emph{JSON}
% \footnote{Further described in \hyperref[appendix:ciujson]{Appendix B}.}
specification containing both structural definitions and keyword-based \textit{tags}. 

In Stage 1, RegEx-based tag matching identifies likely command types to accommodate multiple actions or embedded intents. For example, the command type \texttt{CLIMB\_TO} may be tagged with keywords such as \texttt{climb}, \texttt{altitude}, and \texttt{increase}, allowing a command like \textit{Singapore 123, climb five thousand feet} to be accurately identified. In Stage 2, only filtered candidate structures are passed, where \emph{DSPy}~\citep{khattab2024dspy}, in conjunction with frontier LLMs, extracts precise parameters, conditions, values, and contextual information.

\begin{figure}[h!]
    \includegraphics[width=\linewidth]{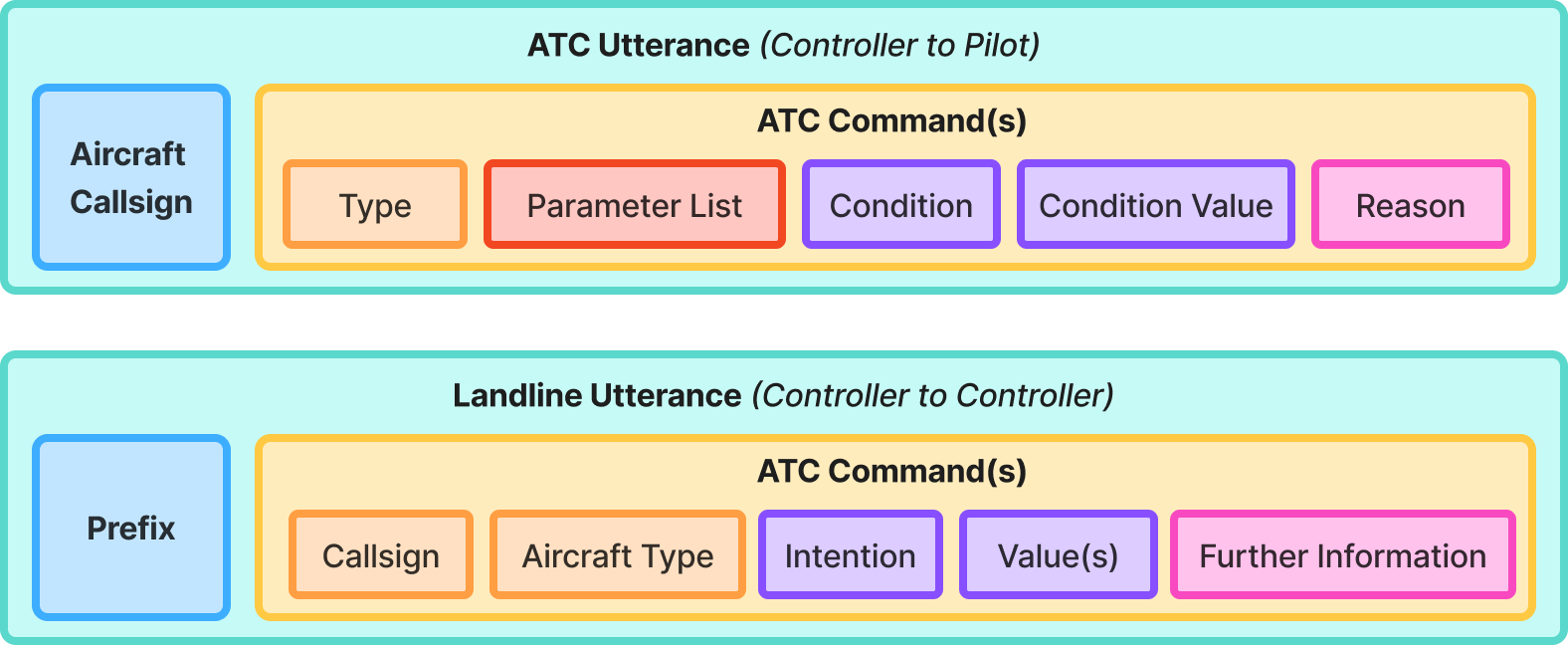}
    \caption{Elements of an ATCO Utterance}
    \label{fig:atc-command}
\end{figure}

As shown in \Cref{fig:atc-command}, each communication type has a differing structural ontology, linguistic patterns and operational constraints
% \footnote{A list of command types and their parameters for controller-to-controller and controller-to-pilot communications is provided in \hyperref[appendix:ppcommandtype]{Appendix C} and \hyperref[appendix:gccommandtype]{Appendix D}.}
. As such, \app{} implements CIU as two specialized modules tailored to each communication type.

\paragraph{Radio Frequency CIU} For controller-pilot interactions, each utterance consists of an aircraft callsign and one of more ATC commands, each containing a type, parameter list, optional conditions and values, and any operational reason.

\paragraph{Landline CIU} For controller-controller coordination, each utterance includes a coordination prefix and ATC command(s) with, callsign, aircraft type, intended action, associated values, and further contextual information. These are considerably more variable due to the large set of potential information.

\subsubsection{Response Generation}

Output from the CIU module is then passed into the Response Generation module, which leverages \emph{DSPy} in conjunction with frontier LLMs. The Response Generation module is divided into two parallel pipelines: a \textit{Pilot Response Generator} and a \textit{Controller Response Generator}, each of which is prompted with role-appropriate conventions and phraseology.

\subsubsection{Text-to-Speech (TTS)}

\app{} implements a domain-adapted text-to-speech (TTS) module to synthesize responses from \textit{pseudo-pilots} and \textit{ghost controllers}. Unlike general-purpose TTS systems, ATC communication requires (i) strict pronunciation of specialized terminology, (ii) a recognizable Singaporean accent, (iii) fast but intelligible delivery, and (iv) stable long-form synthesis without hallucinations. To address these requirements, \app{} adapts three models:

\begin{itemize}[topsep=0pt,nosep]
    \item \emph{XTTS 2.0}, an autoregressive multilingual voice-cloning model with integrated vocoder.
    \item \emph{CSM} (Conversational Speech Model)\footnote{\url{https://hf.co/sesame/csm-1b}}, a cross-lingual text-audio Transformer supporting contextual and expressive synthesis.
    \item \emph{OrpheusTTS}\footnote{\url{https://hf.co/canopylabs/orpheus-3b-0.1-ft}}, a large unified codec-based TTS model capable of high-fidelity generation.
\end{itemize}

\paragraph{Experimental Setup}
All TTS fine-tuning and inference experiments were conducted on a workstation equipped with two NVIDIA GeForce RTX~4090 GPUs (24\,GB each), an Intel Xeon Silver~4210 CPU, and 128\,GB RAM. Depending on the experiment, models were distributed across both GPUs (e.g., parallel training runs or multi-speaker streaming inference). Mixed-precision training and inference were enabled throughout to reduce memory usage and increase synthesis speed.

\paragraph{Dataset Preparation}
TTS models were trained on a mixed corpus constructed specifically for Singapore-accented ATC speech. The dataset combines (i) real Singaporean-accent speech sourced from public corpora and internal recordings and (ii) synthetic ATC-style speech generated using commercial systems such as  Gemini\footnote{\url{https://ai.google.dev/gemini-api/docs/speech-generation}} and ElevenLabs\footnote{\url{https://elevenlabs.io/app/speech-synthesis/text-to-speech}}. Synthetic examples were used to cover a large range of callsigns, waypoints, runway identifiers, and multi-value instructions that are difficult to obtain from natural recordings.

To reduce overfitting and improve robustness, \app{} applies a set of audio augmentations including speed perturbation, pitch shifting, volume scaling, silence trimming, and time-stretching. These augmentations compensate for the overly clean nature of synthetic speech and limited dataset size while increasing the variability of prosody and timing patterns in the training corpus.

The final training corpus comprised Singaporean-accent conversational speech, mixed Singaporean-accent plus synthetic ATC speech, and internal Singaporean-accented aviation speech, supplemented with European ATC data only for auxiliary comparisons.

\paragraph{Accent Adaptation}
Accent adaptation was necessary due to the lack of native Singaporean-accent ATC data. A multi-stage strategy was used. First, conversational Singaporean-accent recordings were combined with scripted ATC commands designed to emphasize local vowel realizations, syllable-timing patterns, and pitch contours characteristic of Singapore English. Second, synthetic ATC speech generated with Singapore-accent settings was incorporated to expand coverage of domain-specific terminology. Third, augmentations were applied to induce prosodic variability and prevent models from learning overly rigid or monotonic speech patterns.

Parameter-efficient fine-tuning using LoRA~\citep{hu2022lora} was applied to XTTS, CSM, and Orpheus. This allowed the models to acquire Singaporean-accent phonetic cues and ATC timing conventions while preserving their multilingual generalization abilities. Multiple datasets (single-speaker, multi-speaker, conversational, and ATC-specific) were evaluated to examine the impact of speaker consistency on accent stability and pronunciation accuracy.

\paragraph{Terminology and Text Processing}
ATC communication contains specialized terminology such as callsigns, waypoints, runway identifiers, and ICAO-standard phrasing. Because these items are largely absent from public speech corpora, \app{} expands terminology coverage through a combined text-normalization and synthetic data construction process:
\begin{itemize}
    \item Custom text-normalization rules that convert transcripts into ICAO-compliant spoken forms (for example digit-by-digit numbers, flight levels, callsign expansions, and runway labels).
    \item Synthetic creation of missing or rare aviation terms using \emph{Gemini} and \emph{ElevenLabs}, including callsigns, waypoint names, and local aerodromes.
\end{itemize}

\paragraph{Training Hyperparameters}
\emph{XTTS} was trained using the Coqui framework, while \emph{CSM} and \emph{Orpheus} were fine-tuned using the Unsloth framework. Across all models, we used the AdamW optimiser, a linear warm-up over the first 1--3\% of steps, followed by cosine learning-rate decay. 

Batch sizes were constrained by model size. \emph{Orpheus} and \emph{CSM} were trained with global batch sizes of 16–24 using gradient accumulation, whereas XTTS used an effective batch size of 32. Gradient clipping at 1.0 was applied throughout to prevent divergence during the early fine-tuning phase.

All models were trained for approximately 10–20 epochs, depending on convergence on the Singapore-accent ATC dataset. This range was chosen to balance adaptation and overfitting risks, given the limited volume of domain-specific data. Validation loss was tracked after every 100 steps, and the checkpoint with minimum validation loss was selected for evaluation. Across runs, loss curves stabilized after roughly 8–12 epochs, with diminishing returns beyond that point.

\paragraph{Generation and Streaming}
\app{} generates real-time TTS audio via chunk-based synthesis where each instruction is split into small, overlapping segments generated sequentially, allowing for parallelization (as implemented for \emph{XTTS}). These partial audio segments are streamed to the frontend immediately via \emph{WebSockets}, reducing end-to-end latency and allowing for parallelization.

Timestamps returned by the backend are used to align chunks and reconstruct a continuous stream, with short fades at chunk boundaries to avoid clicks or abrupt transitions.

\subsection{Simulation of Aircraft Movement (SAM)}

\app{} operates at a fixed polling rate $\Delta t$, simulating radar sweeps occurring at regular intervals. Each aircraft is defined by a set of performance parameters (e.g., climb rate, cruise speed, ceiling, and runway requirements), which are used to accurately replicate real-world dynamics. At each interval, the SAM module computes the aircraft state, represented by the current state vector $S_t$ and its target trajectory $T_t$, defined in Eq. \eqref{eq:state_vector} and \eqref{eq:target_vector} respectively:

\small

% \[
% S_t = \begin{bmatrix}
% \phi_t \emph{ (latitude)} \\
% \lambda_t \emph{ (longitude)} \\
% h_t \emph{ (altitude)} \\
% \psi_t \emph{ (heading)} \\
% V_{\text{IAS,}t} \emph{ (IAS)} \\
% V_{s,t} \emph{ (vertical speed)} \\
% \theta_t \emph{ (bank angle)}
% \end{bmatrix} \quad
% \]

% \[
% T_t = \begin{bmatrix}
% h_{\text{target,}t} \emph{ (target altitude)} \\
% \psi_{\text{target,}t} \emph{ (target heading)} \\
% V_{\text{IAS,target,}t} \emph{ (target IAS)} \\
% \delta_{\text{turn,}t} \emph{ (turn rate)}
% \end{bmatrix}
% \]

\begin{equation}
S_t = \begin{bmatrix}
\phi_t \text{ (latitude)} \\
\lambda_t \text{ (longitude)} \\
h_t \text{ (altitude)} \\
\psi_t \text{ (heading)} \\
V_{\text{IAS,}t} \text{ (IAS)} \\
V_{s,t} \text{ (vertical speed)} \\
\theta_t \text{ (bank angle)}
\end{bmatrix}
\label{eq:state_vector}
\end{equation}

\begin{equation}
T_t = \begin{bmatrix}
h_{\text{target,}t} \text{ (target altitude)} \\
\psi_{\text{target,}t} \text{ (target heading)} \\
V_{\text{IAS,target,}t} \text{ (target IAS)} \\
\delta_{\text{turn,}t} \text{ (turn rate)}
\end{bmatrix}
\label{eq:target_vector}
\end{equation}

\normalsize

When STPs are received by the SAM module, it computes a new aircraft state $S_{t+\Delta t}$ using deterministic motion models. This systematic update loop ensures that aircraft follow realistic trajectories while responding dynamically to control instructions or scenario-driven goals. The update process for heading and altitude is governed by the rules described in Eq. \eqref{eq:heading_rate} through \eqref{eq:alt_increment}. \\

% \subsubsection*{Heading}
% \small
% \[
% \dot{\psi}_t = \frac{1091 \tan(\theta_t)}{V_{\text{IAS,}t}} \quad \text{(or constant if } \theta_t = 0)
% \]
% \[
% \Delta \psi_t = \text{sign}\left(\delta_{\text{turn,}t}\right) \cdot \min\left(\left|\dot{\psi}_t \Delta t\right|, \left|\delta_{\text{turn,}t}\right|\right)
% \]

% \normalsize
% \subsubsection*{Altitude}
% \small
% \[
% V_{s,t} = 
% \begin{cases}
% \min\left(\dot{h}_\text{climb}, \frac{h_{\text{target,}t} - h_t}{\Delta t}\right) & h_{\text{target,}t} \geq h_t \\
% -\min\left(\dot{h}_\text{descent}, \frac{h_t - h_{\text{target,}t}}{\Delta t}\right) & h_{\text{target,}t} < h_t
% \end{cases}
% \]
% \[
% \Delta h_t = V_{s, t} \Delta t
% \]

\subsubsection*{Heading}
\small
\begin{equation}
\dot{\psi}_t = \frac{1091 \tan(\theta_t)}{V_{\text{IAS,}t}} \quad \text{(or constant if } \theta_t = 0)
\label{eq:heading_rate}
\end{equation}
\begin{equation}
\Delta \psi_t = \text{sign}\left(\delta_{\text{turn,}t}\right) \cdot \min\left(\left|\dot{\psi}_t \Delta t\right|, \left|\delta_{\text{turn,}t}\right|\right)
\label{eq:heading_inc}
\end{equation}
\normalsize

\subsubsection*{Altitude}
\small
\begin{equation}
V_{s,t} = 
\begin{cases}
\min\left(\dot{h}_\text{climb}, \frac{h_{\text{target,}t} - h_t}{\Delta t}\right) & h_{\text{target,}t} \geq h_t \\
-\min\left(\dot{h}_\text{descent}, \frac{h_t - h_{\text{target,}t}}{\Delta t}\right) & h_{\text{target,}t} < h_t
\end{cases}
\label{eq:vs_logic}
\end{equation}
\begin{equation}
\Delta h_t = V_{s, t} \Delta t
\label{eq:alt_increment}
\end{equation}

\normalsize
where $h_\text{climb}$ and $h_\text{descent}$ are predefined parameters.
% ~\\This systematic update loop ensures that aircraft follow realistic trajectories while responding dynamically to control instructions or scenario-driven goals.

\subsection{AI-Assisted Radiotelephony Performance Evaluation}
% While existing ATCO training simulators focus primarily on traffic simulation and communication generation, performance evaluation remains largely manual and dependent on instructor expertise. This introduces variability in grading standards and consistency.

% To address this, 
\app{} incorporates an AI-assisted performance evaluation framework designed to standardize and scale the assessment of trainee radiotelephony (RT) communications. The framework analyzes commands at the utterance level to generate objective quantitative scores and actionable qualitative feedback for after-action reviews.

\paragraph{Evaluation Scope and Design Principles}

The proposed evaluation framework focuses on aspects of ATCO performance that can be objectively inferred from radiotelephony communications and simulator state:
\begin{itemize}
    \item \textit{Communication Correctness}: adherence to ICAO-standard phraseology and structure
    \item \textit{Operational Validity}: alignment of instructions with safety constraints and flight trajectories
\end{itemize}

Higher-order competencies (e.g., planning ability, conflict resolution) are outside the current scope, as they require temporal reasoning and direct observation of controller behavior beyond RT transcripts. 

\paragraph{Evaluation Workflow}

Each trainee command is processed through a hybrid architecture combining deterministic rules and LLMs (\Cref{fig:evaluation-pipeline}):

\begin{figure}[h!]
    \centering
    \includegraphics[width=\linewidth]{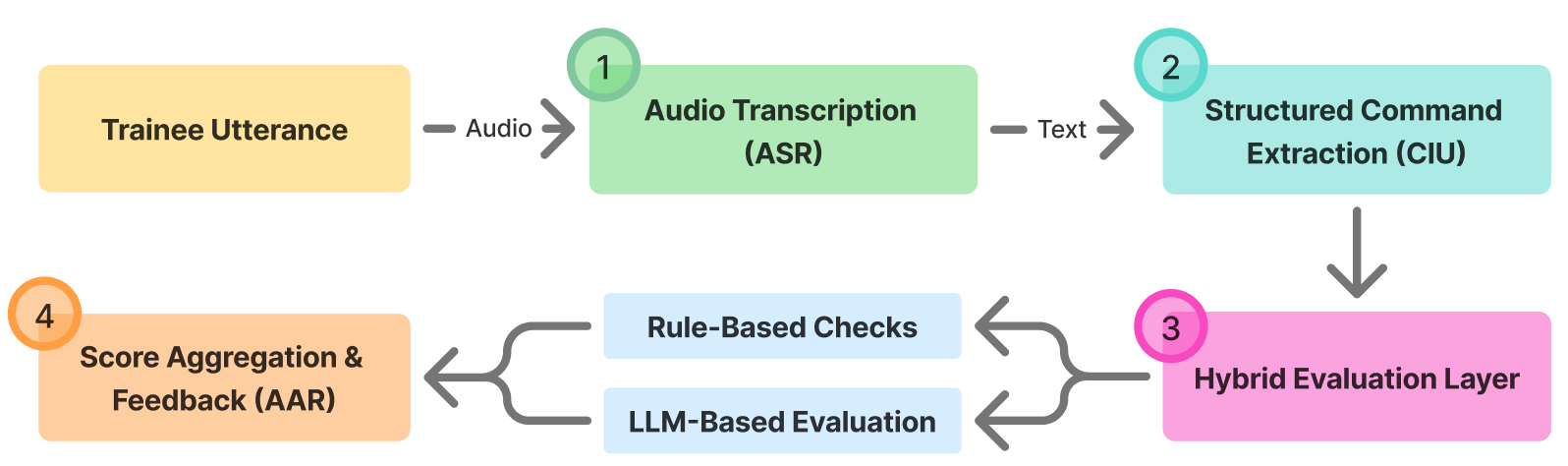}
    \caption{Evaluation pipeline}
    \label{fig:evaluation-pipeline}
\end{figure}

\begin{enumerate}
    \item \textit{Speech Transcription}: Trainee speech is transcribed into text using the ASR module
    \item \textit{Structured Extraction}: The CIU module extracts STPs, including command type, values, and contextual attributes
    \item \textit{Hybrid Evaluation}: The extracted command is evaluated using two complementary components:
        \begin{itemize}
            \item \textit{Rule-Based Evaluation}: Deterministic rules assess compliance with ICAO phraseology, structure, and numerical correctness
            \item \textit{LLM-Based Analysis}: A DSPy-optimized LLM evaluates semantic correctness, contextual intent, and linguistic nuances not captured by rules
        \end{itemize}
    \item \textit{Score Aggregation and Feedback Generation}: Outputs from both components are combined into metric-specific scores and explanatory feedback for after-action review (AAR)
\end{enumerate}

% For example, a trainee command such as \textit{"Descend to 3000 feet and turn heading 180"} is evaluated for phraseology correctness, parameter completeness, and operational validity against the current simulation state.

To enhance robustness, BERT-based contextual embeddings are used to measure semantic similarity between trainee and expected phraseology. These representations capture meaning beyond exact keyword matching, enabling the system to recognize correct intent even when phrasing varies.

This semantic similarity layer complements the DSPy-optimized LLM analysis by providing a structured and consistent similarity signal, reducing over-reliance on generative reasoning alone and improving evaluation stability.

This hybrid approach ensures both \textit{precision} (through rule enforcement) and \textit{flexibility} (through contextual reasoning), across highly structured scenarios and semi-open ATC scenarios where task objectives are fixed but variations in phraseology and response formulation are permitted.

\begin{table}[H]
\centering
\footnotesize
\renewcommand{\arraystretch}{1.1}
\begin{tabular}{lp{4.2cm}}
\toprule
\textbf{Metric} & \textbf{Description} \\
\midrule
Accuracy & Correct use of standard phraseology and values \\
Brevity & Concise transmission, free of redundancy \\
Completeness & Includes all necessary elements \\
Safety Adherence & Maintain safe separation and avoid conflicts \\
Route Compliance & Follows assigned trajectory \\
\bottomrule
\end{tabular}
\vspace{0.5\baselineskip}
\caption{Core performance metrics used in \app{} evaluation framework}
\label{tab:metrics}
\end{table}

\paragraph{Performance Metrics}
\app{} evaluates trainee performance across five core metrics, each scored independently on a scale of 0-100.

\paragraph{Communication Metrics}
The first three metrics evaluate the linguistic and structural quality of RT communications.

\begin{itemize}
    \item \textit{Accuracy}: Measures adherence to ICAO phraseology and correctness of operational values. Rule-based penalties are applied for incorrect terminology and critical safety-impacting errors, while the LLM identifies subtle deviations in phrasing and intent.
    \item \textit{Brevity}: Evaluates whether transmissions are concise and free of redundancy. This includes analysis of word count deviation, speaking rate, filled pauses, and repeated information. Semantic similarity checks ensure that concise commands preserve intended meaning.
    \item \textit{Completeness}: Assesses whether all required parameters for a given command type are present and correctly structured. A slot-based validation mechanism verifies the presence and ordering of mandatory elements based on the intent, while LLMs identify context-dependent omissions.
\end{itemize}

\paragraph{Operational Metrics}

In addition to communication quality, \app{} evaluates whether the instructions issued are operationally valid within the simulation environment.

\begin{itemize}
    \item \textit{Safety Adherence}: Measures whether commands maintain required aircraft separation minima based on aviation safety standards. Two primary criteria are considered: vertical separation ($\ge$ 500 ft) and horizontal separation ($\ge$ 3 NM).
    
    A separation violation is detected when either minimum is not satisfied. Separation events are evaluated across aircraft pairs, and a configurable \textit{time-to-correct (TTC)} window allows trainees to resolve conflicts before penalties are applied. Penalties are assigned based on severity (near-miss, major breach, critical breach), with unresolved violations resulting in greater score reductions.
    \item \textit{Route Compliance}: Evaluates whether aircraft trajectories remain consistent with assigned routes and scenario constraints. Deviations such as missed waypoints and off-track movement are detected using configurable tolerance thresholds. Similarly, a TTC window will also be applied to allow corrective instructions before penalties are enforced.
\end{itemize}

\paragraph{Hybrid Scoring Strategy}

The evaluation framework employs a penalty-based scoring system, where deviations from expected behavior reduce the metric score according to severity. Rule-based penalties ensure consistent handling of deterministic errors, while LLM-generated feedback provides contextual explanations and identifies nuanced issues.

LLM outputs are constrained using structured prompting via DSPy, enabling consistent output formats and reducing variability across evaluations.

The final session score is computed  in Eq. \eqref{eq:session_score} as a weighted aggregation of these metrics, where safety and correctness metrics are assigned higher importance to reflect operational priorities:

\begin{equation}
S = 100 - \sum_{i=1}^{n} w_i P_i
\label{eq:session_score}
\end{equation}

where \( P_i \) represents penalties incurred from individual deviations, and \( w_i \) denotes the corresponding weight assigned to each metric.   

\paragraph{Session-Level Feedback and Replay System}

Beyond per-command evaluation, \app{} aggregates results across a training session to generate a comprehensive post-session performance report. This includes:

\begin{itemize}
    \item Timeline of all trainee commands
    \item Individual metric score breakdown
    \item Identified strengths and errors
    \item Natural language feedback for improvement
\end{itemize}

To support detailed analysis, the system integrates synchronized \textit{audio playback} and \textit{visual replay} of aircraft movements. This allows trainees and instructors to review decisions in context, bridging communication performance with operational outcomes.

By combining structured rule-based validation, BERT-based semantic similarity, and LLM-driven contextual analysis, the proposed framework bridges the gap between \textit{linguistic correctness} and \textit{operational intent}. This enables consistent, explainable, and scalable evaluation of ATCO radiotelephony performance, reducing reliance on subjective instructor judgment while enhancing training quality.

\begin{table*}[t]
    \centering
    \small
    \begin{tabular}{lccc|cc|cc}
        \toprule
        & & \multicolumn{2}{c|}{\thead{ATCOSIM}}
        & \multicolumn{2}{c|}{\thead{MNSC ASR Part 1}}
        & \multicolumn{2}{c}{\thead{SG-Aviation}} \\
        
        \thead{Model} & \thead{Params}
        & WER & CER
        & WER & CER
        & WER & CER \\
        \midrule
        
        \whisper{} (Large) & 1.5B
        & 80.49 & 53.81
        & 24.44 & \cellcolor{yellow!20}7.07
        & 78.81 & 35.68 \\
        
        \whisper{} (Small) & 0.2B
        & \cellcolor{red!20}91.05 & \cellcolor{red!20}61.51
        & 27.05 & 8.67
        & \cellcolor{red!20}107.80 & \cellcolor{red!20}80.57 \\
        
        \emph{Parakeet} & 0.6B
        & 53.22 & 23.07
        & 21.90 & \cellcolor{green!20}6.07
        & 63.12 & 25.83 \\
        
        \emph{WhisperATC} (Large) & 1.5B
        & \cellcolor{green!20}22.5 & \cellcolor{green!20}12.17
        & \cellcolor{red!20}35.61 & \cellcolor{red!20}21.24
        & \cellcolor{yellow!20}51.16 & \cellcolor{yellow!20}24.33 \\
        
        \emph{WhisperATC} (Small) & 0.2B
        & 54.33 & 19.79
        & \cellcolor{green!20}18.68 & 7.34
        & \cellcolor{green!20}40.93 & \cellcolor{green!20}16.52 \\
        
        \emph{MERaLiON 2} & 10B
        & \cellcolor{yellow!20}48.63 & \cellcolor{yellow!20}19.56
        & \cellcolor{yellow!20}19.48 & 11.44
        & 73.95 & 36.11 \\
        \midrule
        
        \whisper{} (Fine-tuned) & 1.5B
        & 21.67 & 13.25
        & 21.17 & 17.43
        & 19.08 & 15.51 \\
        
        \app{} ASR Pipeline (\Cref{fig:asr-pipeline}) & 1.5B
        & 20.72 & 14.14
        & 19.63 & 15.19
        & 14.45 & 11.62 \\
        \bottomrule
    \end{tabular}
    \vspace{0.5\baselineskip}
    \caption{ASR performance across datasets (WER/CER)}
    \label{tab:asr-scores}
\end{table*}

%%%%%%%%%%%%%%%%%%%%%%%%%%%%%%%%%%%%%%%%%%%%%%%%%%%%%%%%%%%%%%%%%%%%%%%%%%%%%%%%%%%%%%%%%%%%%%%%%%%%%%%%%%%%%%%%%%%%%%%%%%
% EXPERIMENTAL RESULTS
%%%%%%%%%%%%%%%%%%%%%%%%%%%%%%%%%%%%%%%%%%%%%%%%%%%%%%%%%%%%%%%%%%%%%%%%%%%%%%%%%%%%%%%%%%%%%%%%%%%%%%%%%%%%%%%%%%%%%%%%%%
\section{Experimental Results} \label{results}

To assess the effectiveness of the proposed methods, we conducted a series of evaluations designed to measure their performance that leveraged task-appropriate qualitative and quantitative indicators. 

\subsection{Automatic Speech Recognition (ASR)}

The performance of several ASR models was evaluated on the following public and internal datasets using word error rate (WER) and character error rate (CER):
\begin{itemize}
    \item ATCOSIM~\citep{hofbauer2008atcosim}, which predominantly collected Western ATCO speech
    \item Multi-Task National Speech Corpus (MNSC)~\citep{wang2025advancing} ASR Part 1, which collected Singaporean-accented speech
    \item SG-Aviation (internal dataset), which collected Singaporean-accented ATCO speech
    % \item Gen-SG-Aviation (internal dataset), which ...
\end{itemize}

As seen in \Cref{tab:asr-scores}, base \whisper{} models perform poorly across all datasets, indicating the need for accent- and domain-specific adaptation. \emph{WhisperATC} Small remains the strongest evaluated model for Singaporean-accented speech, while \emph{MERaLiON} achieves competitive results on ATCOSIM and MNSC but degrades on SG-Aviation. 

Based on these evaluations, \whisper{} Large v3 was selected as the base model and fine-tuned on a combined corpus of ATCOSIM, Singaporean-accented ATC recordings, and augmented synthetic data. The fine-tuned model and full ASR pipeline (\Cref{fig:asr-pipeline}) achieve the lowest SG-Aviation WER of 26.08\% and 23.45\% respectively. Error rates on the out-of-domain ATCOSIM and MNSC datasets are higher, showing the trade-off of domain-specific adaptation. While specialized models exhibit high variance across datasets, the \app{} pipeline maintains a consistent error profile within the 20--35\% range, showing its stability across both Western and local aviation contexts.

\subsection{Text-to-Speech (TTS)}

TTS models were evaluated for real-time ATC simulation using perceptual quality via Mean Opinion Scores (MOS) and A/B comparisons. Audio clips were anonymized as \emph{Model A}, \emph{B}, and \emph{C} to avoid recognition bias.

% \afterpage{%
%     \clearpage% Flush earlier floats (otherwise order might not be correct)
%     \thispagestyle{empty}% empty page style (?)
%     \begin{landscape}% Landscape page

% \centering
% \begin{adjustwidth}{0pt}{24pt}
		% \begin{flushleft}

\begin{table*}[ht]
    \centering
    \footnotesize
    \setlength{\tabcolsep}{4pt}
    \begin{tabular}{llcccccccccc}
        \toprule
        \thead{Model} & \thead{Source}
        & Clarity & Pron. & Prosody & Naturalness & Overall
        & {Gender Acc.} & Male & Female & {\# Ratings} \\
        \midrule
        \textsc{XTTS}   & Human
        & \cellcolor{green!20}4.06 & \cellcolor{yellow!20}3.81 & \cellcolor{yellow!20}3.74 & \cellcolor{green!20}3.79 & \cellcolor{green!20}3.71
        & \cellcolor{yellow!20}98.41 & 31 & 32 & 63 \\
        
        \textsc{CSM}    & Human
        & \cellcolor{red!20}3.52 & \cellcolor{red!20}3.50 & \cellcolor{red!20}3.27 & \cellcolor{red!20}3.44 & \cellcolor{red!20}3.21
        & \cellcolor{green!20}100.00 & 35 & 27 & 62 \\
        
        \textsc{Orpheus} & Human
        & \cellcolor{yellow!20}3.95 & \cellcolor{green!20}3.87 & \cellcolor{green!20}3.77 & \cellcolor{yellow!20}3.54 & \cellcolor{yellow!20}3.70
        & \cellcolor{red!20}96.83 & 34 & 29 & 63 \\
        \midrule
        \textsc{XTTS}   & LLM
        & \cellcolor{yellow!20}4.13 & \cellcolor{yellow!20}4.57 & \cellcolor{red!20}3.36 & \cellcolor{red!20}3.07 & \cellcolor{red!20}3.37
        & \cellcolor{red!20}92.68 & 205 & 205 & 410 \\
        
        \textsc{CSM}    & LLM
        & \cellcolor{red!20}4.10 & \cellcolor{yellow!20}4.57 & \cellcolor{yellow!20}3.41 & \cellcolor{yellow!20}3.10 & \cellcolor{yellow!20}3.44
        & \cellcolor{yellow!20}95.15 & 205 & 207 & 412 \\
        
        \textsc{Orpheus} & LLM
        & \cellcolor{green!20}4.22 & \cellcolor{green!20}4.67 & \cellcolor{green!20}3.51 & \cellcolor{green!20}3.19 & \cellcolor{green!20}3.59
        & \cellcolor{green!20}95.60 & 205 & 204 & 409 \\
        \bottomrule
    \end{tabular}
    \vspace{0.5\baselineskip}
    \caption{Comparison of human and LLM-based MOS scores for TTS models.}
    \label{tab:tts-mos}
\end{table*}

\begin{table*}[ht]
    \centering
    % \footnotesize
    \begin{tabular}{lcccc}
        \toprule
        \thead{Comparison} 
        & \thead{Option 1 (\%)} 
        & \thead{Option 2 (\%)} 
        & \thead{No Pref. (\%)} 
        & \thead{\# Ratings} \\
        \midrule
        XTTS vs CSM        & \cellcolor{green!20}43.3 & \cellcolor{red!20}30.0 & 26.7 & 30 \\
        XTTS vs Orpheus    & \cellcolor{green!20}44.9 & \cellcolor{yellow!20}34.7 & 20.4 & 49 \\
        CSM vs Orpheus     & \cellcolor{red!20}34.0 & \cellcolor{yellow!20}53.2 & 12.8 & 47 \\
        \midrule
        Male vs Female (XTTS)    & 35.7 & \cellcolor{green!20}40.5 & 23.8 & 42 \\
        Male vs Female (CSM)     & 19.4 & \cellcolor{red!20}63.9 & 16.7 & 36 \\
        Male vs Female (Orpheus) & 22.2 & \cellcolor{yellow!20}44.4 & 33.3 & 36 \\
        \bottomrule
    \end{tabular}
    \vspace{0.5\baselineskip}
    \caption{A/B preference test results for model-level and gender-level comparisons.}
    \label{tab:tts-ab}
\end{table*}

% \end{flushleft}
% \end{adjustwidth}

%     \end{landscape}
%     \clearpage% Flush page
% }

\paragraph{Mean Opinion Scores (MOS)}
MOS was rated on five dimensions using male and female voices on a diverse set of ATC utterances. Ratings were provided by  
(i) 21 aviation personnel familiar with ICAO radiotelephony, and  
(ii) an automated LLM evaluator (\emph{Gemini 2.5 Flash})  
to provide robustness and highlight divergences between human and model perception. Both followed brief guidelines on a (1--5 scale, 5 being optimal):

\begin{enumerate}[topsep=0pt,nosep]
    \item \textbf{Clarity}: intelligibility of all spoken content
    \item \textbf{Pronunciation}: correctness of terminologies
    \item \textbf{Prosody}: naturalness of rhythm and pacing
    \item \textbf{Naturalness}: closeness to human speech
    \item \textbf{Overall Quality}: overall quality, reflecting other dimensions and any artifacts
\end{enumerate}

Each human rater evaluated 9 audio clips covering all models and both genders, with optional annotations for gender mismatch or comments.

As shown in \Cref{tab:tts-mos}, \emph{XTTS} and \emph{Orpheus} achieve the highest overall human MOS scores (3.71 and 3.70 respectively). \emph{XTTS} leads on clarity, making it well-suited for intelligibility-critical ATC communications, while \emph{Orpheus} scores higher on prosody and pronunciation, producing more natural-sounding delivery. \emph{CSM} trails both models across all dimensions, largely due to truncations and incomplete utterance endings that disrupted perceived quality.

LLM-based scores follow the same ranking, with \emph{Orpheus} scoring highest at 3.59. Notably, \emph{XTTS} exhibits the largest human--LLM divergence (3.71 vs.\ 3.37), suggesting that the automated evaluator underweights the prosodic and acoustic qualities that human raters — particularly those accustomed to ATC speech — find most salient. Gender accuracy is high across all three models, exceeding 96\% in every case.

\paragraph{A/B Comparison Tests}
Pairwise A/B tests were conducted with human evaluators to capture relative preferences. Two comparison types were used:
(i) \textit{Model A/B Comparisons}: pairs of clips generated from the same script and gender,
enabling direct comparison of model behavior.
(ii) \textit{Gender Comparisons}: pairs comparing male and female voices produced by the same model.

Raters selected their preferred clip for each pair and could optionally provide a brief justification.

\Cref{tab:tts-ab} shows a clear preference for both \emph{XTTS} and \emph{Orpheus} over \emph{CSM}. \emph{XTTS} is favored over \emph{CSM} 43.3\% vs 30.0\%, and \emph{Orpheus} shows an even stronger advantage at 53.2\% vs 34.0\%, aligning with \emph{CSM}’s lower MOS scores and frequent truncations. Between \emph{XTTS} and \emph{Orpheus}, \emph{XTTS} is preferred (44.9\% vs 34.7\%) for clarity, while \emph{Orpheus} is noted for smoother pacing. In gender comparisons, female voices are preferred across all models, with comments noting smoother and more natural delivery for the female voices of \emph{XTTS} and \emph{Orpheus}.

\subsection{AI-Assisted Radiotelephony Performance Evaluation}

To evaluate the effectiveness of the proposed communication assessment module, we conducted experiments on three key communication factors: \textit{accuracy}, \textit{brevity}, and \textit{completeness}. Each evaluator was optimized using the DSPy framework to compare various strategies, specifically \textit{BootstrapFewShotWithRandomSearch} (hereafter referred to as RS) and \textit{MIPROv2}. These methods leverage a teacher model to bootstrap high-quality demonstrations; however, MIPROv2 distinguishes itself by jointly optimizing the underlying task instructions along these demonstrations.

All evaluators were trained on labeled examples and evaluated on held-out development sets. Performance was measured using inverted Mean Absolute Error (MAE), where higher scores indicate better alignment with ground truth grading. For brevity, BERTScore (F1) was additionally used to capture semantic similarity between generated and expected radiotelephony, ensuring both linguistic correctness and scoring consistency.

\begin{table*}[h]
\centering
\begin{tabular}{lccc}
\toprule
\textbf{Evaluator} & \textbf{Baseline (\%)} & \textbf{Compiled (\%)} & \textbf{Improvement} \\
\midrule
Accuracy     & 83.8 & 91.7 & +7.9 \\
Brevity      & 85.5 & 89.7 & +4.2 \\
Completeness & 81.5 & 88.1 & +6.6 \\
\bottomrule
\end{tabular}
\vspace{0.5\baselineskip}
\caption{Performance of Communication Evaluators Before and After DSPy Optimization}
\label{tab:communication-results}
\end{table*}
\vspace{-0.5\baselineskip}

\Cref{tab:communication-results} shows prompt optimization improved performance across all evaluators, with the largest gain observed in \textit{accuracy} (+7.9\%).

\begin{table*}[h]
\centering
\begin{tabular}{lccc}
\toprule
\textbf{Evaluator} & \textbf{FewShot (\%)} & \textbf{FewShot+RS (\%)} & \textbf{MIPROv2 (\%)} \\
\midrule
Accuracy     & 83.8 & \textbf{91.7} & 88.2 \\
Brevity      & 85.5 & 88.2          & \textbf{89.7} \\
Completeness & 81.5 & 86.9          & \textbf{88.1} \\
\bottomrule
\end{tabular}
\caption{Comparison of DSPy Optimization Strategies across Communication Factors}
\label{tab:optimization-benchmarks}
\end{table*}
\vspace{-0.5\baselineskip}

To further evaluate the impact of different optimization strategies, \Cref{tab:optimization-benchmarks} compares standard \textit{BootstrapFewShot} with \textit{RandomSearch} (RS) and \textit{MIPROv2}. While RS achieved the peak score for \textit{accuracy}, \textit{MIPROv2} yielded the best results for \textit{brevity} and \textit{completeness}. This suggests that instruction-aware optimization is particularly beneficial for stylistic and multi-parameter assessments, whereas random search excels in calibrating purely phraseology-based scoring.

%%%%%%%%%%%%%%%%%%%%%%%%%%%%%%%%%%%%%%%%%%%%%%%%%%%%%%%%%%%%%%%%%%%%%%%%%%%%%%%%%%%%%%%%%%%%%%%%%%%%%%%%%%%%%%%%%%%%%%%%%%
% FUTURE WORK
%%%%%%%%%%%%%%%%%%%%%%%%%%%%%%%%%%%%%%%%%%%%%%%%%%%%%%%%%%%%%%%%%%%%%%%%%%%%%%%%%%%%%%%%%%%%%%%%%%%%%%%%%%%%%%%%%%%%%%%%%%
\section{Limitations \& Future Work} \label{futurework}

While \app{} aims to provide a realistic ATCO training simulator with an in-built end-to-end communications pipeline, we acknowledge some limitations that will be addressed as feature enhancements in future work.

\paragraph{ASR Performance} The fine-tuned model is trained exclusively on Singapore ATC data, limiting generalization to other airspaces, accents, and languages. Future work will expand the training corpus with international ATC recordings, radio static profiles, and filler words ("uh", "um") to improve noise robustness and filler detection accuracy for performance evaluation. Techniques such as \textit{contextual biasing} via \emph{TCPGen}~\citep{sun2023tcpgen} and Keyword Spotting will also be explored to further improve recognition of domain-specific terminology.

\paragraph{Live Scenario Injects} \app{} supports a set of predefined scenarios. Future work will implement the functionality for instructors to inject live edits to the scenarios, allowing it to be more customized and challenging.

% \paragraph{Live Performance Evaluation} We aim to leverage the live communications and simulator data from \app{} to evaluate ATCO performance in a controlled setting.

\paragraph{TTS Hallucinations} Long text inputs caused our TTS models to produce errors like jumbled phrasing and cut-offs. Noting that autoregressive TTS models are unstable on long sequences and that chunk-wise decoding reduces alignment drift~\citep{li2025robustefficientautoregressivespeech}, we plan to apply inference-time text chunking as a simple, practical first step, with more complex methods considered only if needed.

\paragraph{TTS Evaluation Constraints} The human MOS evaluation was conducted with 21 raters, each evaluating 9 clips — a sample size that limits the statistical robustness of the results. Future evaluations should recruit a larger and more diverse pool of aviation personnel to improve confidence in perceptual quality rankings. The automated LLM evaluator provides useful signal at scale but diverges from human judgement on prosodic and acoustic dimensions, as seen in the XTTS human--LLM gap in \Cref{tab:tts-mos}. Calibrating such evaluators against a larger human baseline remains an important direction for future work.

\paragraph{Adaptive Scenario Generation} Rather than relying solely on fixed scenarios, \app{} supports the automatic generation of training situations based on trainee performance. By analyzing real-time interactions and RT performance metrics, the system identifies performance gaps and dynamically introduces scenarios that target specific weaknesses. For example, a trainee demonstrating difficulty with altitude management may be presented with scenarios of increased altitude-control complexity. This creates a closed-loop training environment that continuously adapts to the trainee’s competency, reinforcing learning without requiring direct instructor intervention.

%%%%%%%%%%%%%%%%%%%%%%%%%%%%%%%%%%%%%%%%%%%%%%%%%%%%%%%%%%%%%%%%%%%%%%%%%%%%%%%%%%%%%%%%%%%%%%%%%%%%%%%%%%%%%%%%%%%%%%%%%%
% CONCLUSION
%%%%%%%%%%%%%%%%%%%%%%%%%%%%%%%%%%%%%%%%%%%%%%%%%%%%%%%%%%%%%%%%%%%%%%%%%%%%%%%%%%%%%%%%%%%%%%%%%%%%%%%%%%%%%%%%%%%%%%%%%%
\section{Conclusion} \label{conclusion}
% In this paper, we propose \app{}, a training simulator enabling autonomous ATCO training without instructors by combining an end-to-end speech pipeline with advanced speech recognition, custom text-to-speech models, and systems for parsing and responding to ATC utterances.

In this paper, we propose \app{}, a training simulator enabling autonomous ATCO training without \textit{simpilots} by combining an end-to-end speech pipeline with advanced speech recognition, custom text-to-speech models, and systems for parsing and responding to ATC utterances. Designed for the Singaporean context, \app{} addresses reliance on simpilots and localization limitations, enabling scalable, high-fidelity, instructor-independent training.

% commented out for submission
\section*{Acknowledgment} \label{acknowledgements}
This paper is made possible with the support of RSAF Agile Innovation Digital (RAiD), Republic of Singapore Air Force, Singapore.

% \afterpage{%
%     \clearpage% Flush earlier floats (otherwise order might not be correct)
%     \thispagestyle{empty}% empty page style (?)
%     \begin{landscape}% Landscape page

% \begin{table*}[ht]

% \end{table*}

%     \end{landscape}
%     \clearpage% Flush page
% }

\bibliography{references}

% \clearpage

% \appendix

% \section*{Appendix}

% \startappendixtoc        % 这句一定要在 appendix 的 section 出现前
% \setcounter{tocdepth}{2} % 控制 appendix 目录深度（可选）

% \appendixtableofcontents % 打印 Appendix Contents（只会包含 appendix）

\end{document}